\documentclass{article} 
\usepackage{iclr2026_conference,times}

\usepackage{amsmath,amsfonts,bm}

\def\eqref#1{equation~\ref{#1}}

\def\1{\bm{1}}

\DeclareMathAlphabet{\mathsfit}{\encodingdefault}{\sfdefault}{m}{sl}
\SetMathAlphabet{\mathsfit}{bold}{\encodingdefault}{\sfdefault}{bx}{n}

\usepackage{hyperref}
\usepackage{url}
\usepackage{enumitem}
\usepackage{booktabs}
\usepackage{amsmath, amsthm, amssymb}
\usepackage{amsmath} 
\usepackage{amsfonts} 
\usepackage{courier}
\usepackage{xcolor}
\usepackage[table]{xcolor}
\usepackage{colortbl}
\definecolor{LightGreen1}{rgb}{0.90, 1.0, 0.90} 
\definecolor{LightGreen2}{rgb}{0.75, 0.95, 0.75} 
\definecolor{LightGreen3}{rgb}{0.60, 0.90, 0.60} 
\definecolor{LightGreen4}{rgb}{0.45, 0.85, 0.45} 
\definecolor{LightRed1}{rgb}{1.0, 0.90, 0.90} 
\definecolor{LightRed2}{rgb}{0.95, 0.75, 0.75} 
\definecolor{LightWhite}{rgb}{1.0, 1.0, 1.0} 
\usepackage{array}
\usepackage[utf8]{inputenc}
\usepackage[T1]{fontenc}
\usepackage{hyperref}
\usepackage{url}
\usepackage{booktabs}
\usepackage{amsfonts}
\usepackage{nicefrac}
\usepackage{microtype}
\usepackage{xcolor}
\usepackage{graphicx}
\usepackage{lipsum}

\usepackage{listings}
\usepackage{xcolor}

\usepackage[table]{xcolor}
\usepackage{pgf} 

\definecolor{perfGreen}{HTML}{C7E9C0}
\definecolor{perfYellow}{HTML}{FFE27A} 
\definecolor{perfRed}{HTML}{F5A3A3}    

\newcommand{\THgood}{0.80} 
\newcommand{\THwarn}{0.50} 

\newcommand{\scorecell}[2]{%
  \begingroup
  \pgfmathparse{#1}%
  \edef\_val{\pgfmathresult}%
  \ifdim \_val pt < \THwarn pt
    \cellcolor{perfRed}#2%
  \else\ifdim \_val pt < \THgood pt
    \cellcolor{perfYellow}#2%
  \else
    \cellcolor{perfGreen}#2%
  \fi\fi
  \endgroup
}

\newcommand{\chip}[2]{%
  \begingroup
  \pgfmathparse{#1/#2}\edef\val{\pgfmathresult}%
  \def\bg{perfGreen}%
  \ifdim \val pt < \THwarn pt \def\bg{perfRed}%
  \else\ifdim \val pt < \THgood pt \def\bg{perfYellow}\fi\fi
  \colorbox{\bg}{\kern0.5pt {\tiny \rule{0pt}{0.72ex}\rule[-0.10ex]{0pt}{0pt}$#1/#2$}\kern0.5pt}%
  \endgroup
}

\newcommand{\scorecellpct}[1]{\scorecell{(#1)/100}{#1}}

\newcommand{\single}[1]{\scorecell{#1}{#1}}

\usepackage{multirow}
\usepackage{algorithm}
\usepackage{algpseudocode}
\usepackage{subcaption}  
\usepackage{nicematrix}

\usepackage{tcolorbox}
\usepackage{xcolor}
\tcbuselibrary{theorems}
\usepackage{amsthm}

\usepackage{graphicx}
\usepackage{wrapfig}

\title{FedAgentBench: Towards Automating Real-world Federated Medical Image Analysis with Server–Client LLM Agents}

\author{
Pramit Saha$^{1}$, Joshua Strong$^{1}$, Divyanshu Mishra$^{1}$, Cheng Ouyang$^{1*}$, J.~Alison Noble$^{1*}$ \\
$^{1}$Department of Engineering Science, University of Oxford \\
{\tt\small pramit.saha@eng.ox.ac.uk} \\
\small $^{*}$Equal supervision
}

\iclrfinalcopy 
\begin{document}

\maketitle
\begin{abstract}
Federated learning (FL) allows collaborative model training across healthcare sites without sharing sensitive patient data. However, real-world FL deployment is often hindered by complex operational challenges that demand substantial human efforts in cross-client coordination and data engineering. This includes: (a) selecting appropriate clients (hospitals), (b) coordinating between the central server and clients, (c) client-level data pre-processing, (d) harmonizing non-standardized data and labels across clients, and (e) selecting FL algorithms based on user instructions and cross-client data characteristics. However, the existing FL works overlook these practical orchestration challenges. These operational bottlenecks motivate the need for autonomous, agent-driven FL systems, where intelligent agents at each hospital client and the central server agent collaboratively manage FL setup and model training with minimal human intervention. To this end, we first introduce: (i) an agent-driven FL framework that captures key phases of real-world FL workflows from client selection to training completion, and (ii) a benchmark dubbed FedAgentBench that evaluates the ability of LLM agents to autonomously coordinate healthcare FL.
Our framework incorporates 40 FL algorithms, each tailored to address diverse task-specific requirements and cross-client characteristics. Furthermore, we introduce a diverse set of complex tasks across 201 carefully curated datasets, simulating 6 modality-specific real-world healthcare environments, \textit{viz.}, Dermatoscopy, Ultrasound, Fundus, Histopathology, MRI, and X-Ray. We assess the agentic performance of 14 open-source and 10 proprietary LLMs spanning small, medium, and large model scales. While some agent cores such as GPT-4.1 and DeepSeek V3 can automate various stages of the FL pipeline, our results reveal that more complex, interdependent tasks based on implicit goals remain challenging for even the strongest models. 
\end{abstract}

\section{Introduction and Background}

Federated Learning (FL) \cite{li2021survey,mcmahan2017communication,li2020federated} allows collaborative model training across multiple healthcare institutions (\textit{e.g.}, hospitals) without sharing raw medical data. 
A typical FL workflow involves several tightly coupled components: selecting suitable clients for training, preprocessing heterogeneous data locally, harmonizing labels and datasets across clients, coordinating client-server communication, selecting optimal FL algorithm, and aggregating model updates in the server. These components must be executed in a precise and orchestrated manner across multiple clients.
Real-world execution of an FL pipeline necessitates close coordination by human data scientists and machine learning engineers in server and client locations, who are tasked with managing a range of demanding communicational and technical operations. These include selecting appropriate client nodes based on task relevance and resource availability, implementing local data preprocessing pipelines (e.g., normalization, filtering, schema mapping), and harmonizing cross-site inconsistencies of data and label spaces. Additionally, they must determine the most suitable FL algorithms, and manage training schedules and aggregation strategies. This manual orchestration poses a significant barrier to scalable and real-time deployment of FL, particularly in sensitive domains like healthcare, where institutions store diverse yet complementary datasets that cannot be centralized due to privacy and compliance constraints.
Moreover, many healthcare facilities, especially in low- and middle-income countries (LMICs) and rural areas, lack the resources to hire dedicated data scientists or machine learning engineers, further limiting their ability to participate in FL initiatives despite having valuable local data. To this end, in this paper, we investigate the capabilities of LLM Agents in tackling these issues with minimal human intervention.

\begin{figure*}[t]
    \centering
\includegraphics[width=1\columnwidth]{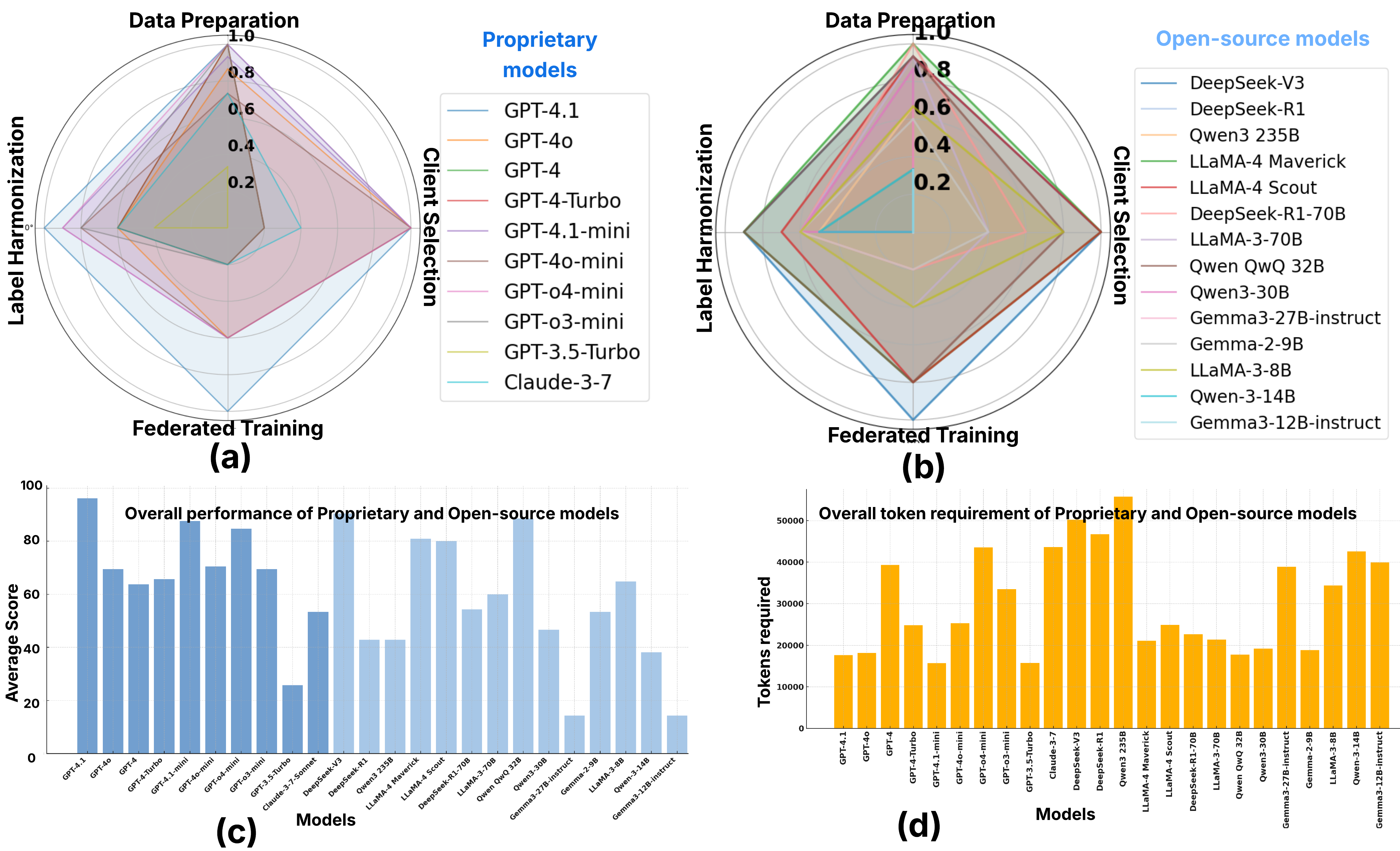}
    \caption{Performance of 24 LLM Agents on 4 FL sub-tasks over 6 healthcare environments. (a) and (b) show the performance of proprietary and open-source models respectively on four subtasks each, \textit{viz.}, Client Selection, Data preprocessing, Label Harmonization, and Federated Training. (c) and (d) show the average score and mean overall token requirement of all models across all tasks.} 
\label{fig:11}
\end{figure*}

The rapid advancement of LLMs has led to the emergence of autonomous AI agents capable of executing complex, multi-step tasks across various domains \cite{gurreal,goutora,cailarge,li2023camel,wang2023voyager,wu2024autogen,mei2024aios,chu2025llm,qiu2024llm,luo2025large}. This capability can be particularly transformative for real-world healthcare FL, where agent-based automation can reduce the operational burden on healthcare sites and enable broader participation in collaborative AI development. There are no existing works on agent-driven FL workflow; for general-purpose agents or agentic FL works, refer to \textbf{Related Works in Appendix~A}.

To this end, we introduce an agentic FL framework (see Figs. 2 \& 3) along with a benchmark \textbf{FedAgentBench} (see Fig. 1), designed to systematically evaluate the performance of LLM-driven agents in orchestrating FL workflows. To ensure comprehensive coverage, we incorporate 201 datasets, 6 major medical imaging modalities, and 40 representative FL algorithms designed for diverse real-world healthcare objectives and cross-client data compositions. To the best of our knowledge, this is the first work addressing FL problem-solving capabilities of LLM Agents directly dealing with server and client interactions. Our benchmark makes the following key contributions:

\textbf{(1) Technical contribution:} We first \textbf{present a plug-and-play modular agentic FL framework} supporting 40 FL algorithms and 24 LLM agents. It also allows for easy integration of new FL algorithms, agents and tasks with minimal adaptation. It is a unified FL framework with multi-faceted scenarios, multiple imaging modalities, and complex FL workflow structures. It encompasses four realistic and interlinked agent-driven FL phases: (i) \textbf{Client Selection}, where server and client agents communicate dataset suitability, (ii) \textbf{Data Preprocessing}, involving data restructuring, cleaning, and standardization using learned tools, (iii) \textbf{Label Harmonization}, where agents align inconsistent label taxonomies across clients, and (iv) \textbf{Federated Model Training}, where selected algorithms are deployed in a decentralized setup. It is worth noting that while we simulate healthcare environments in this work, the framework can be readily extended to other FL settings such as finance, IoT, etc.

\textbf{(2) Dataset and Task contribution:} To evaluate the effectiveness of LLM agents in real-world healthcare tasks, we construct a realistic simulation of inter-hospital collaboration within a FL framework in representative clinical scenarios. Specifically, we curate and publicly release \textbf{six medical imaging FL agentic environments} comprising a total of \textbf{201 datasets} and a diverse collection of tasks spanning a range of difficulties. To introduce greater variability across clients, we systematically modify the original image resolutions, file format extensions, and intensity distributions of the client datasets. Additionally, we carefully inject noisy and irrelevant samples spanning images from other modalities, text files,  and other extraneous formats into client data directories to simulate realistic uncurated data environments and reflect the challenges of real-world clinical settings. 

\textbf{(3) Empirical contribution:} 
As a part of FedAgentBench, we evaluate the performance of 24 LLM agents across diverse FL tasks based on task completion rate (\textit{i.e.}, success rate), token efficiency, and overall time required. We investigate how varying levels of prompt granularity affect task execution and systematically compare agent performance across different autonomy tiers: guided tool invocation, autonomous planning, and fully independent script generation. Our analysis provides a comprehensive assessment of agentic capabilities and limitations in supporting real-world collaborative healthcare workflows. We will open-source and continuously update the benchmark on Github repository to support FL research and help healthcare data holders fully realize the value of cross-silo data.

\begin{figure*}
    \centering
\includegraphics[width=1\columnwidth]{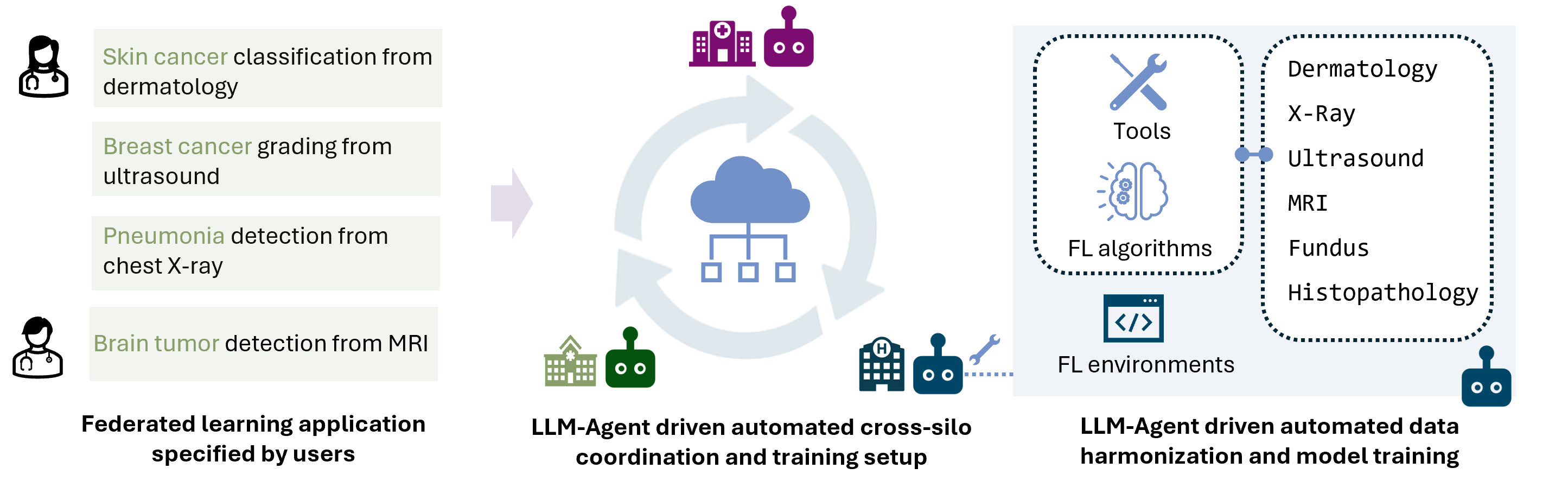}
    \caption{Overview of our agent-driven FL setup. First, user defines task specification. Accordingly, LLM agents perform server-client coordination and complete required tasks using available tools and FL algorithms in any of the 6 modality-specific healthcare environments.} 
\label{fig:11}
\end{figure*}

\section{FedAgentBench Framework}
\subsection{Problem Formulation and Overview}
Given a user-defined task specification for federated medical image analysis, denoted as $\mathcal{T}$, our objective is to construct and execute a complete FL pipeline through collaborative decision-making by a set of autonomous agents. As outlined in Fig. 3, {FedAgentBench} consists of two main components: \textbf{(i) Federated medical imaging workspace $\mathcal{W}$} which can be sub-categorized to server workspace $\mathcal{W}_s$ and client workspace $\mathcal{W}_c$ as well as \textbf{(ii) Multi-agent coordination system $\mathcal{A}$}. 
The workspace $\mathcal{W}$ encapsulates the critical resources required for FL pipeline construction and includes: (1) client metadata files (data cards) containing natural language descriptions of local datasets (in $\mathcal{W}_c$), (2) FL algorithm specifications (in $\mathcal{W}_s$) and tool usage descriptions (in $\mathcal{W}_c$ and $\mathcal{W}_s$) and (3) structured code templates for each phase of the FL workflow (in $\mathcal{W}_c$ and $\mathcal{W}_s$). 

Built on top of this workspace, the agents operate under a divide-and-conquer strategy to address the complexity and modularity of the entire FL process. The server-client agent system $\mathcal{A} = \{S_1, S_2, S_3, S_4, C_1, C_2, C_3\}$ comprises 7 role-specialized LLM agents (see Fig. 3) responsible for: (1) client selection and server-client communication or orchestration ($S_1, S_2, C_1$), (2) data preprocessing and cleaning ($C_2$), (3) label harmonization ($C_3$), and (4) federated model selection and training ($S_3, S_4$). The collaborative pipeline proceeds iteratively as agents can invoke tools, write scripts, or reason over workspace content to solve subtasks, with execution feedback enabling adaptation. This process can be formally represented as: $\{D_i, R_i\} = \mathcal{A}(D_{i-1}, R_{i-1}, \mathcal{T} \mid \mathcal{W})$ where $D_i$ denotes the code, decisions, or configurations generated or modified in the $i$-th iteration, and $R_i$ represents execution results or tool feedback (e.g., logs, errors, evaluation metrics), with $D_0 = R_0 = \emptyset$.
The goal is to produce a complete, executable FL pipeline satisfying task specification $\mathcal{T}$, measured in terms of success and efficiency under real-world constraints simulated by $\mathcal{W}$.

\begin{figure*}
    \centering
\includegraphics[width=1\columnwidth]{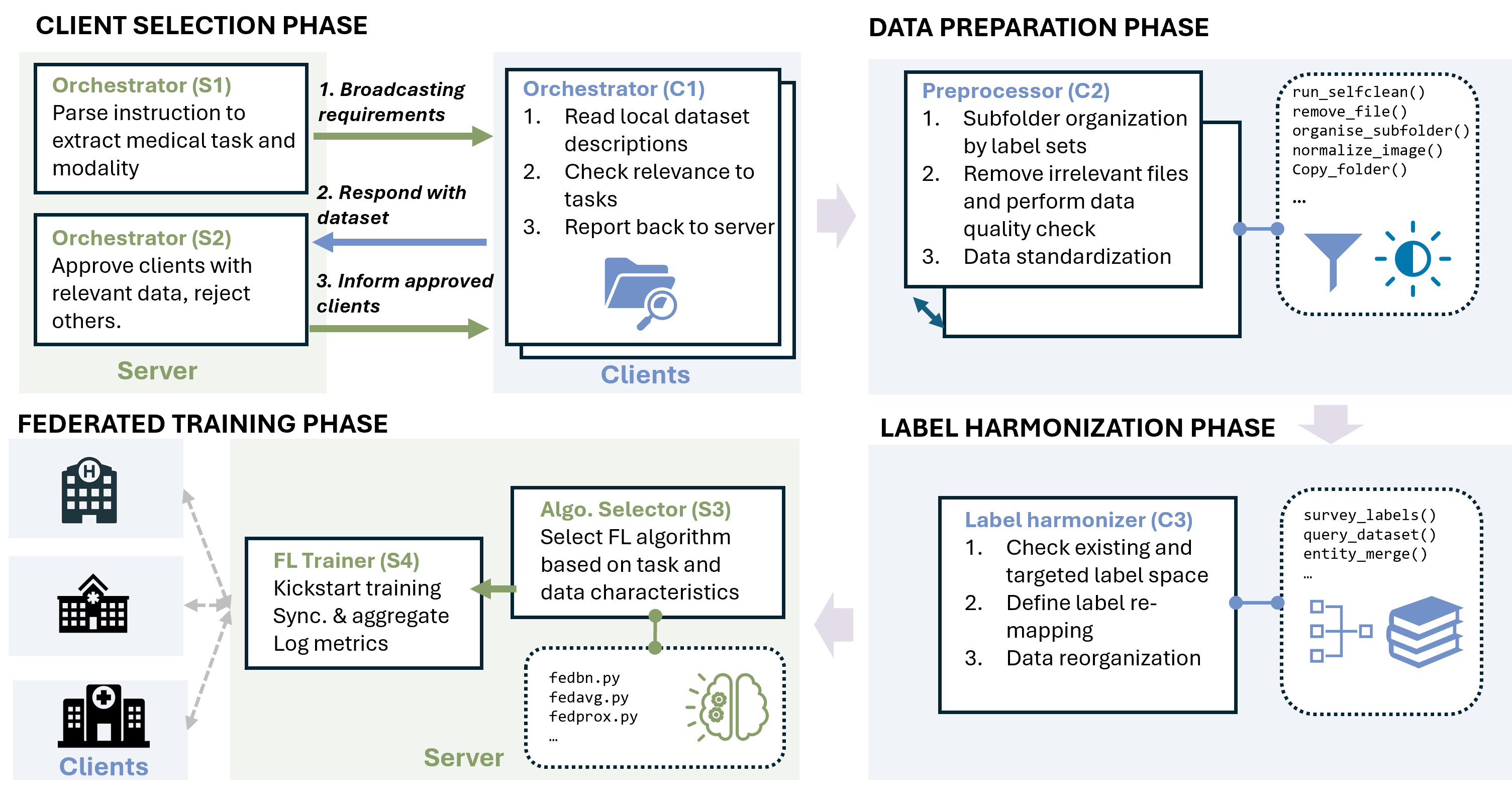}
    \caption{An overview of the FedAgentBench Framework. It comprises 7 role-specialized LLM agents $(S_1, S_2, S_3, S_4, C_1, C_2, C_3)$ for completing 4 distinct phases of the FL workflow (see \S 2.3)
} 
\label{fig:11}
\end{figure*}

\subsection{Client Dataset Curation and FL Algorithm Integration}
\noindent \textbf{Broad coverage of real-world medical specialties and data sets:} We construct FedAgentBench clients by adapting \textbf{201 publicly available datasets} with 2D and 3D dimensionality across 6 different medical imaging modalities \textit{viz.} \textbf{25 Dermatology, 33 Ultrasound, 63 Fundus, 32 X-Ray, 28 MRI, and 20 Histopathology datasets}. It spans a broad range of tasks, including disease classification (e.g., tumor detection, cancer subtype identification), disease staging or grading (e.g., cancer and diabetic retinopathy severity levels), anatomical or pathological region segmentation (e.g., tumor or stroke localization), object detection, regression, reconstruction, \textit{etc}. Each client is simulated to comprise one or more of these datasets, reflecting the diversity and heterogeneity typical of real-world healthcare institutions. We construct a datacard accompanying each client based on the metadata sourced from its original publication, repository or website. \textbf{See Appendix C.1 \& Listings 6-8}.


\textbf{Cross-client data heterogeneity beyond distribution shifts:} In order to introduce greater variability across clients and better emulate the heterogeneity found in real-world clinical data silos, we systematically modify several aspects of the original datasets:

\textbf{(i) Structured Dataset Perturbations:} We introduce systematic modifications to dataset characteristics, such as varying image resolutions (e.g., downsampling images), altering file format extensions (e.g., converting \texttt{.png} files to \texttt{.jpeg}, \texttt{.bmp}, or \texttt{.tiff}), and modifying intensity distributions to reflect differences in scanner settings or preprocessing pipelines.\\
\textbf{(ii) Inclusion of Uncurated and Irrelevant Files:} To reflect the messiness of real-world clinical storage, we inject non-image and unrelated files into client directories. These include textual notes (\texttt{.txt}, \texttt{.doc}, \texttt{.pdf}), spreadsheets (\texttt{.csv}, \texttt{.xls}), and miscellaneous files (e.g., \texttt{.log}, \texttt{.xml}, \texttt{.ini}). For example, our dermatoscopy dataset contains lesion images mixed with dermatologist notes in \texttt{.pdf} format and other unrelated documents.\\
\textbf{(iii) Simulation of Label and Modality Noise:} We simulate common data quality issues by introducing random duplication of 2-5 samples, injecting 2-5 anatomically or modality-inconsistent images, and deliberately corrupting labels of 2-5 samples to model annotation noise in each dataset.

These artifacts challenge the robustness of agent-based preprocessing and reflect the complexities encountered in real hospital PACS or data repositories. See Appendix C for more details.

\textbf{Algorithm suite for a wide spectrum of FL settings:} As a part of the benchmark design, we also curate a comprehensive suite of \textbf{40 FL algorithms} by integrating and adapting existing implementations. This algorithm collection spans a broad spectrum of FL paradigms enabling standardized and reproducible evaluation across diverse medical imaging tasks \textbf{(See Appendix \S C.4)}. This includes: 

\textbf{(i) Classical FL algorithms} such as \texttt{FedAvg}, \texttt{FedProx}, and \texttt{Scaffold}, which address baseline aggregation and client drift;
\textbf{(ii) Personalized FL algorithms} like \texttt{Per-FedAvg}, \texttt{pFedMe}, and \texttt{FedRep}, which tailor models to heterogeneous client data distributions; 
\textbf{(iii) Regularization-based approaches} like \texttt{Ditto} which impose constraints to preserve global knowledge during local updates; 
\textbf{(iv) Knowledge Distillation-based methods} such as \texttt{FedDF}, enabling model-agnostic communication via logits;
\textbf{(v) Domain generalization techniques} like \texttt{FedSR}, \texttt{FedDG}, and \texttt{FedIRM}, which aim to learn invariant representations across non-IID clients; and 
\textbf{(vi) Optimization and scheduling variants}, such as \texttt{FedNova} which address stability, and convergence rate.

\subsection{Federated Agentic Framework Construction}

FL workflows typically follow a common set of phases, from which we abstract the key human roles and tasks fundamental to their execution as discussed below \textbf{(See Appendix B.2 for more details)}:

\textbf{{1. Client orchestrator agents}}: 
These agents act as the coordinators of the framework by communicating between the server and clients as well as by selecting the most suitable clients for the task based on the user requirements and individual client responses (see Fig. 4).
\begin{wrapfigure}{r}{.85\textwidth}
  \centering 
  \includegraphics[width=\linewidth]{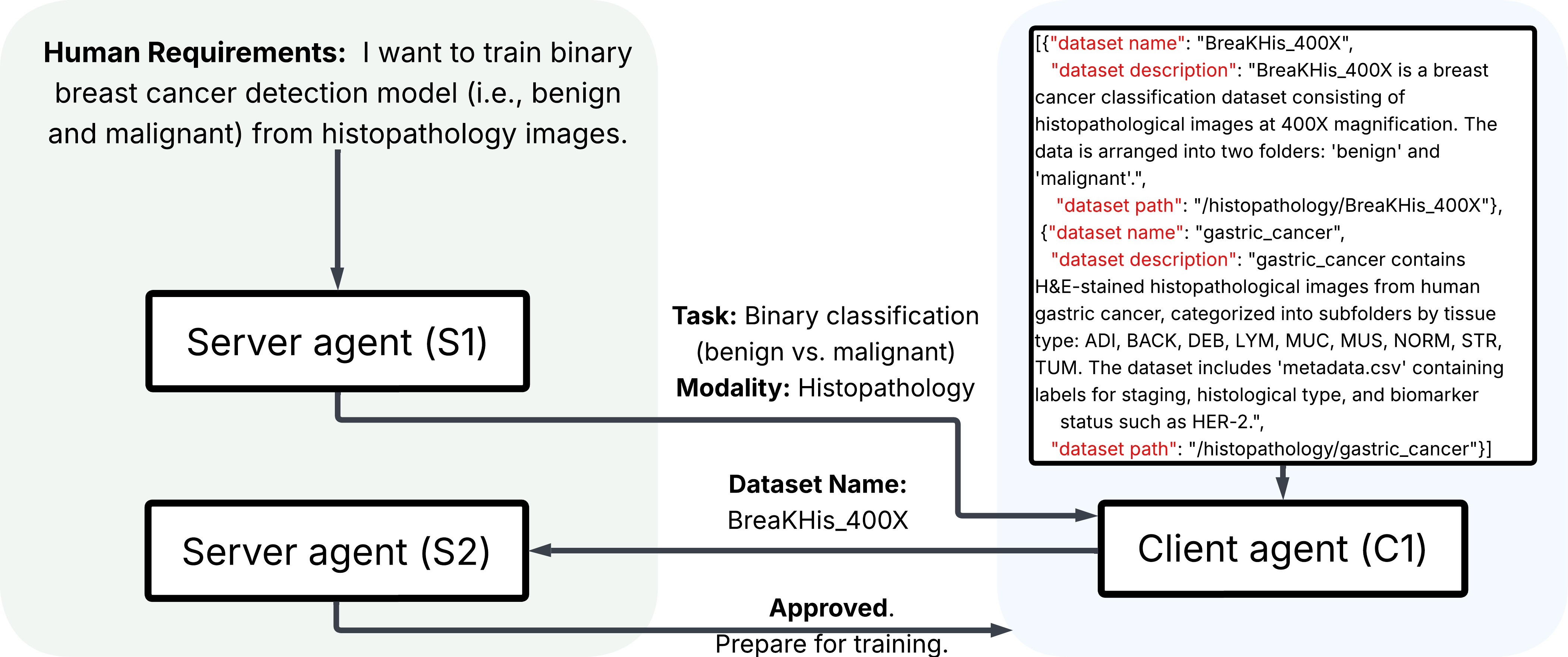} 
    \caption{Client orchestrator agents $S_1$, $C_1$, and $S_2$ in a histopathology-based breast cancer classification task}
    \label{fig:prior_exp}
\end{wrapfigure}
 Server agent $S_1$ interprets the user-defined task $\mathcal{T}$ and communicates imaging modality/task requirements to initiate client selection. For this, it first parses $\mathcal{T}$ and broadcasts a query to all Client Agents (\textit{i.e.}, healthcare sites). Each Client Agent $C_1$ reads local dataset description file, which contains metadata about available datasets, including label sets/imaging types.
Based on semantic and modality matching, $C_1$ evaluates relevance of its datasets to $\mathcal{T}$, returning only matching datasets (if any). Server Agent $S_2$ collects these responses and selects a subset of relevant clients $\mathcal{C}_{\text{sel}}$, which are then approved for further processing \textbf{(see Figs. 9-14 in Appendix D)}.

\textbf{{2. Data pre-processor agent}}: 
It is responsible for preparing selected client datasets for effective participation in the FL pipeline. Given the diversity of data storage formats and quality issues across real-world sites, Data pre-processor agent $C_2$ at each client ensures that the dataset adheres to a standardized structure and meets minimum quality criteria. 
Concretely, it is responsible for standardizing and cleaning datasets at each selected client (see Fig. 5). This includes:

\textbf{(i) Subfolder Organization}: Verifies whether datasets are organized into class-specific subfolders. If disorganized, $C_2$ restructures the folder hierarchy.\\

\textbf{(ii) File Cleaning}: Removes irrelevant files (non-image formats \texttt{.txt}, \texttt{.csv} \textit{etc.}) to ensure format consistency.\\
\textbf{(iii) Data Cleaning}: Detects and flags duplicates, off-topic samples, and noisy labels, which are then removed.
This ensures all selected clients have curated structurally consistent data, enabling downstream harmonization and consequent training \textbf{(see Figs. 34-35 in Appendix D)}.\\ 
\begin{wrapfigure}{r}{.85\textwidth}
  \centering 
  \includegraphics[width=\linewidth]{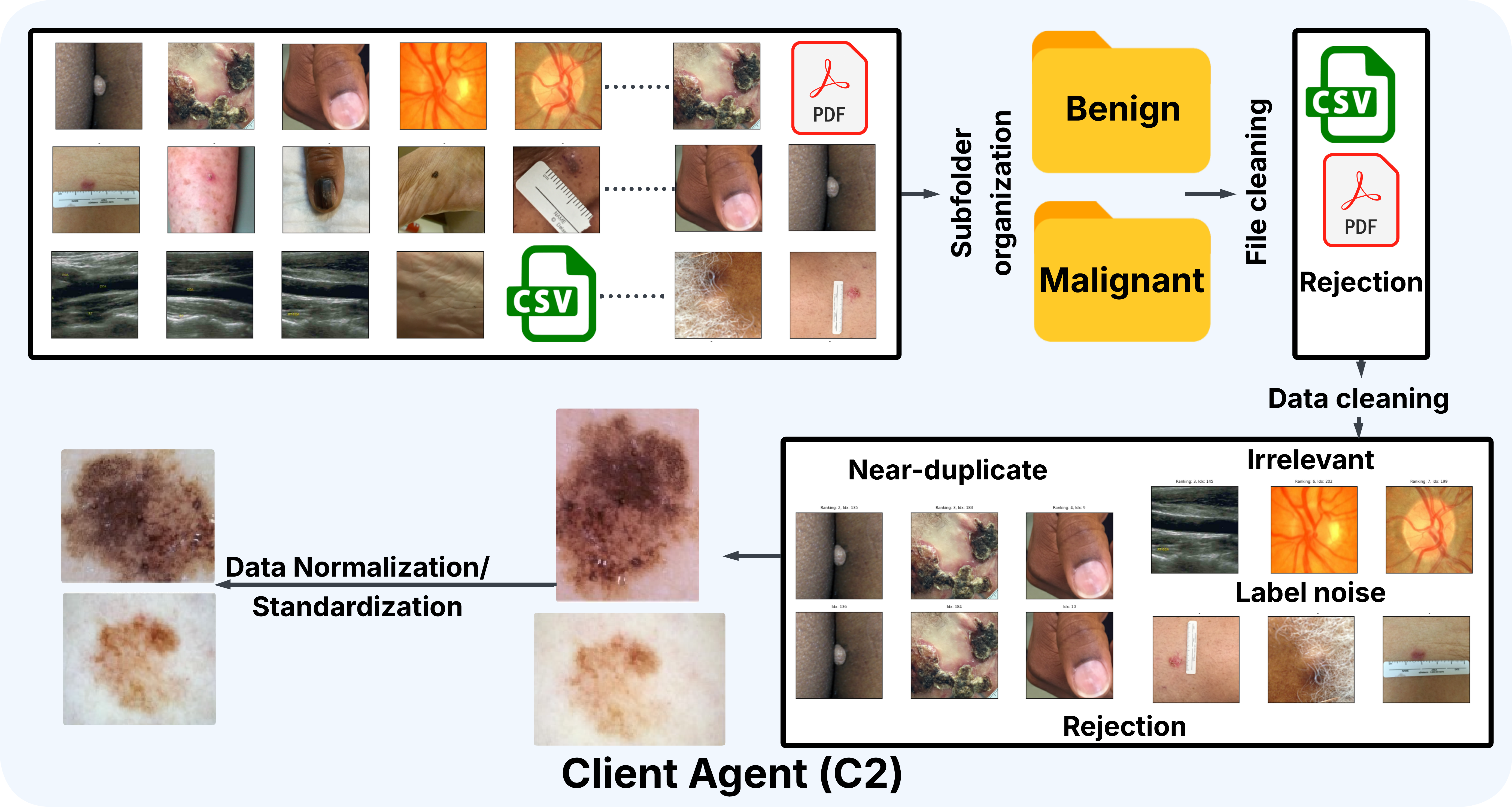} 
    \caption{Data pre-processor agent $C_2$ in skin cancer detection task}
    \label{fig:prior_exp}
\end{wrapfigure}
\textbf{(iv) Data Normalization/Standardization}: Standardizes images across clients based on resolution, intensity, and file extension. This agent thus plays an essential role in bridging the gap between raw, heterogeneous clinical data and the clean, harmonized inputs required for FL. Its operations ensure that all participating clients contribute structurally consistent, high-quality data harmonized across clients, which is crucial for the success of the overall FL system.

\textbf{{3. Task-conditioned label harmonizer agent}}:
This agent ($C_3$) addresses one of the most critical challenges in multi-institutional FL, \textit{ i.e.},  the  inconsistency in label nomenclature and granularity across client datasets (see Fig. 6). 
Due to variations in annotation protocols, terminologies, and domain-specific taxonomies, class labels across clients may not align semantically or structurally. 
 $C_3$ plays a pivotal role in reconciling these differences based on the user requirements: 
\textbf{(i) Class Inspection}: Enumerates all class labels present in client datasets.\\
\textbf{(ii) Label Mapping}: Converts fine-grained labels (e.g., \texttt{"melanoma"}, \texttt{"nevus"}) to harmonized classes (e.g., \texttt{"malignant"}, \texttt{"benign"}) according to a self-developed mapping schema. \\
\begin{wrapfigure}{r}{.85\textwidth}
  \centering 
  \includegraphics[width=\linewidth]{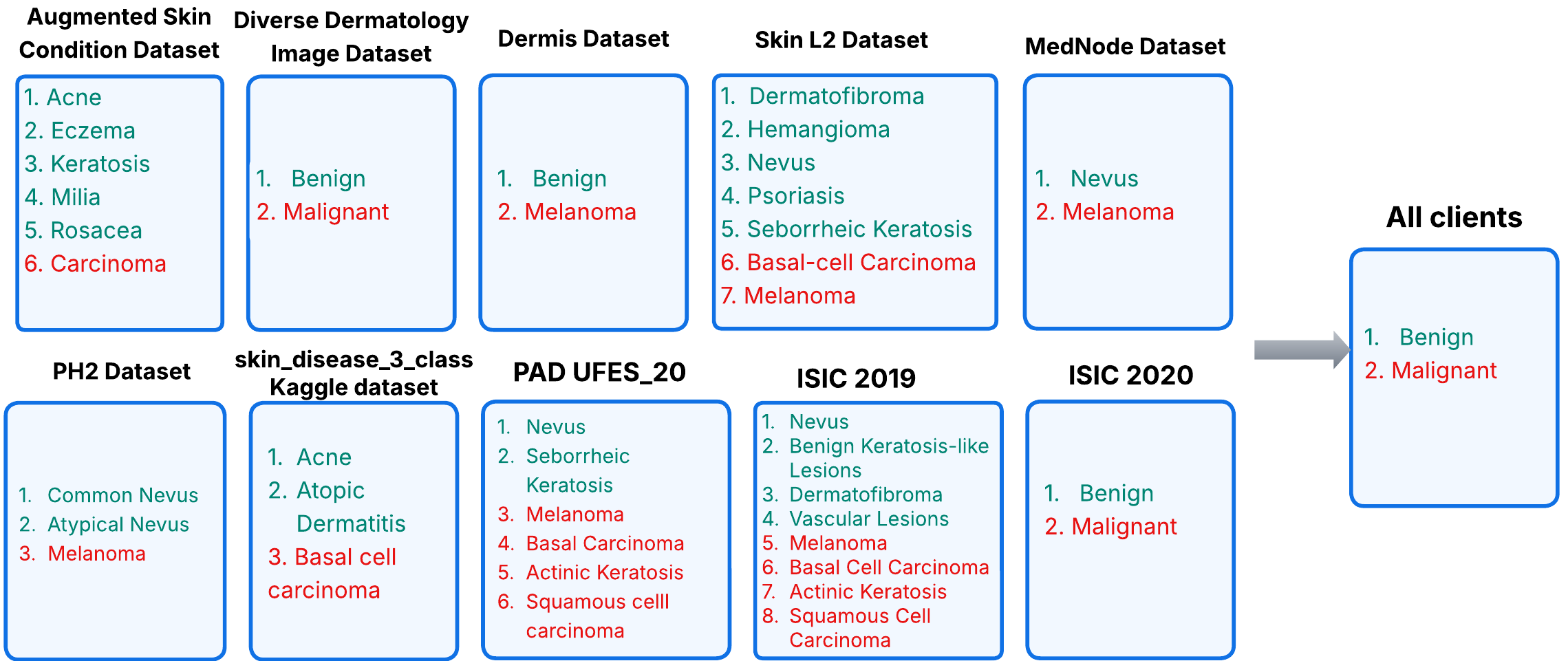} 
    \caption{Label harmonization by agent $C_3$ in dermatology-based skin cancer detection (benign/malignant classes color-coded in green/red)}
    \label{fig:prior_exp}
\end{wrapfigure}
\textbf{(iii) Data Reorganization}: Reorganizes the dataset structure to reflect harmonized labels, aligning image samples with their mapped class definitions. This standardization enables cross-client training without semantic conflicts in label interpretation. 

Through these actions, the agent guarantees that all clients adhere to a shared label vocabulary. 


\textbf{{4. Federated trainer agents:}}
These agents are responsible for initiating the actual federated training process across the selected set of clients and play a central role in converting the prepared environment into a functioning FL system. 
They initiate and coordinate federated training in 2 steps: 

(i) Based on $\mathcal{T}$, \textbf{FL Algorithm Selector Agent} ($S_3$) queries a registry of 40 FL algorithms containing the algorithmic descriptions and then selects a suitable method (e.g., \texttt{FedAvg}, \texttt{pFedSim}, \texttt{FedSR}) based on user requirements. \\
(ii) \textbf{Trainer Agent} ($S_4$) then distributes training details to approved clients and executes Federated Training.
During training, $S_4$ logs per-client and global metrics (e.g., accuracy) and performs model aggregation. Its modular structure supports plug-and-play experimentation with different FL algorithms and training configurations.

\subsection{Privacy Preserving and Modular Design}
A key advantage of our framework is its modular design across phases and agent specializations: Each agent component and phase can be independently evaluated, replaced, or extended. More importantly, this modularity enables future expansion of the benchmark and adaptation to diverse real-world scenarios. For instance, additional components simulating privacy/safety audits conducted by humans or AI can be seamlessly inserted between server and client agents or workflow phases, without the need for altering the existing workflow.

It is to be noted that our framework enforces data privacy by design, aligning fully with FL principles. We explicitly prevent agents from ever accessing or transmitting raw data, model weights, or sensitive metadata. The server receives approvals/configuration signals only, not images, so the agent layer never handles patient data. Instead, agents operate at orchestration layer only and exchange only predefined information (JSON configs, file paths, status signals). They do not have direct access to raw client data (e.g., patient images) or sensitive metadata and never transmit patient data or intermediate outputs externally. Training is invoked via a tool wrapper that runs locally per client; no raw data leaves clients at the agent layer, \textit{i.e.}, federated training is triggered by the agent, but executed on local clients via tools. All data preprocessing and label harmonization also happen locally at clients. Eg: In label harmonization, the agent creates mapping logic, but the mapping execution and label replacement are performed entirely on the local client side.

\begin{table*}[t]
\centering
\caption{Comparison of LLM agents in \textbf{Dermatology} environment based on skin cancer detection task. Here \textbf{P, R, F1} indicate Precision, Recall, and F1 score. \textbf{S, D, F} indicate Schema Compliance Rate, Duplicate Removal Rate, and Format Normalization Rate.\textbf{ E, C, Co} indicate Exact-match Accuracy, Coverage Rate, and Conflict Rate. \textbf{T} indicates Training-start verification score.}
\label{tab:fg-gg-like-fig2}
\resizebox{\textwidth}{!}{
\begin{tabular}{|l|cccc|cccc|}
\toprule
\textbf{Model} & \multicolumn{4}{|c|}{\textbf{Fine-grained guidance}} & \multicolumn{4}{|c|}{\textbf{Goal-oriented guidance}} \\
\hline
  & \textbf{Client-Sel} & \textbf{Data-Pre} & \textbf{Label-Harm} & \textbf{Fed-Train} &  \textbf{Client-Sel} & \textbf{Data-Pre} & \textbf{Label-Harm} & \textbf{Fed-Train}  
  \\ \hline
      & \textbf{P, R, F1} & \textbf{S, D, F} & \textbf{E, C, Co} & \textbf{T}& \textbf{P, R, F1} & \textbf{S, D, F} & \textbf{E, C, Co} & \textbf{T}\\
\hline
\multicolumn{9}{|c|}{\textbf{Proprietary Models}} \\
\midrule
GPT-4.1  &
\scorecell{(0.96+1.00+0.98)/3}{0.96, 1.00, 0.98} &
\scorecell{(1.00+0.97+1.00)/3}{1.00, 0.97, 1.00} &
\scorecell{(0.61+0.65+0.35)/3}{0.61, 0.65, 0.35} &
\single{0.99} &
\scorecell{(0.88+0.86+0.87)/3}{0.88, 0.86, 0.87} &
\scorecell{(1.00+0.96+0.98)/3}{1.00, 0.96, 0.98} &
\scorecell{(0.61+0.61+0.39)/3}{0.61, 0.61, 0.39} &
\single{0.85} \\

GPT-4o  &
\scorecell{(0.88+0.89+0.88)/3}{0.88, 0.89, 0.88} &
\scorecell{(1.00+0.94+0.95)/3}{1.00, 0.94, 0.95} &
\scorecell{(0.18+0.27+0.73)/3}{0.18, 0.27, 0.73} &
\single{0.21} &
\scorecell{(0.79+0.76+0.77)/3}{0.79, 0.76, 0.77} &
\scorecell{(0.96+0.91+0.92)/3}{0.96, 0.91, 0.92} &
\scorecell{(0.16+0.24+0.76)/3}{0.16, 0.24, 0.76} &
\single{0.18} \\

GPT-4  &
\scorecell{(1.00+0.92+0.96)/3}{1.00, 0.92, 0.96} &
\scorecell{(0.02+0.01+0.00)/3}{0.02, 0.01, 0.00} &
\scorecell{(0.22+0.29+0.71)/3}{0.22, 0.29, 0.71} &
\single{0.61} &
\scorecell{(0.70+0.68+0.69)/3}{0.70, 0.68, 0.69} &
\scorecell{(0.05+0.00+0.00)/3}{0.05, 0.00, 0.00} &
\scorecell{(0.00+0.01+0.96)/3}{0.00, 0.01, 0.96} &
\single{0.43} \\

GPT-4-Turbo  &
\scorecell{(0.91+0.89+0.90)/3}{0.91, 0.89, 0.90} &
\scorecell{(0.41+0.33+0.39)/3}{0.41, 0.33, 0.39} &
\scorecell{(0.19+0.24+0.76)/3}{0.19, 0.24, 0.76} &
\single{0.64} &
\scorecell{(0.88+0.79+0.83)/3}{0.88, 0.79, 0.83} &
\scorecell{(1.00+0.98+0.97)/3}{1.00, 0.98, 0.97} &
\scorecell{(0.25+0.29+0.71)/3}{0.25, 0.29, 0.71} &
\single{0.45} \\

GPT-4.1-mini  &
\scorecell{(1.00+1.00+1.00)/3}{1.00, 1.00, 1.00} &
\scorecell{(1.00+0.93+0.98)/3}{1.00, 0.93, 0.98} &
\scorecell{(0.59+0.65+0.35)/3}{0.59, 0.65, 0.35} &
\single{0.61} &
\scorecell{(1.00+0.97+0.98)/3}{1.00, 0.97, 0.98} &
\scorecell{(0.57+0.53+0.57)/3}{0.57, 0.53, 0.57} &
\scorecell{(0.59+0.60+0.40)/3}{0.59, 0.60, 0.40} &
\single{0.58} \\

GPT-4o-mini  &
\scorecell{(0.64+0.61+0.62)/3}{0.64, 0.61, 0.62} &
\scorecell{(1.00+0.92+1.00)/3}{1.00, 0.92, 1.00} &
\scorecell{(0.60+0.63+0.37)/3}{0.60, 0.63, 0.37} &
\single{0.61} &
\scorecell{(0.50+0.56+0.53)/3}{0.50, 0.56, 0.53} &
\scorecell{(1.00+0.96+0.98)/3}{1.00, 0.96, 0.98} &
\scorecell{(0.23+0.26+0.74)/3}{0.23, 0.26, 0.74} &
\single{0.40} \\

GPT-o4-mini  &
\scorecell{(0.94+0.91+0.92)/3}{0.94, 0.91, 0.92} &
\scorecell{(0.98+0.95+0.96)/3}{0.98, 0.95, 0.96} &
\scorecell{(0.63+0.71+0.29)/3}{0.63, 0.71, 0.29} &
\single{0.57} &
\scorecell{(0.90+0.80+0.85)/3}{0.90, 0.80, 0.85} &
\scorecell{(0.74+0.70+0.73)/3}{0.74, 0.70, 0.73} &
\scorecell{(0.45+0.50+0.50)/3}{0.45, 0.50, 0.50} &
\single{0.60} \\

GPT-o3-mini  &
\scorecell{(0.86+0.89+0.87)/3}{0.86, 0.89, 0.87} &
\scorecell{(0.00+0.00+0.00)/3}{0.00, 0.00, 0.00} &
\scorecell{(0.45+0.49+0.51)/3}{0.45, 0.49, 0.51} &
\single{0.58} &
\scorecell{(0.71+0.77+0.74)/3}{0.71, 0.77, 0.74} &
\scorecell{(0.05+0.00+0.00)/3}{0.05, 0.00, 0.00} &
\scorecell{(0.44+0.50+0.50)/3}{0.44, 0.50, 0.50} &
\single{0.63} \\

GPT-3.5-Turbo  &
\scorecell{(0.32+0.35+0.33)/3}{0.32, 0.35, 0.33} &
\scorecell{(0.04+0.00+0.00)/3}{0.04, 0.00, 0.00} &
\scorecell{(0.00+0.03+0.97)/3}{0.00, 0.03, 0.97} &
\single{0.18} &
\scorecell{(0.41+0.30+0.35)/3}{0.41, 0.30, 0.35} &
\scorecell{(0.43+0.38+0.38)/3}{0.43, 0.38, 0.38} &
\scorecell{(0.00+0.00+1.00)/3}{0.00, 0.00, 1.00} &
\single{0.21} \\

Claude-3-7-Sonnet  &
\scorecell{(0.67+0.68+0.67)/3}{0.67, 0.68, 0.67} &
\scorecell{(0.44+0.42+0.42)/3}{0.44, 0.42, 0.42} &
\scorecell{(0.21+0.27+0.73)/3}{0.21, 0.27, 0.73} &
\single{0.42} &
\scorecell{(0.69+0.69+0.69)/3}{0.69, 0.69, 0.69} &
\scorecell{(0.40+0.38+0.39)/3}{0.40, 0.38, 0.39} &
\scorecell{(0.26+0.32+0.68)/3}{0.26, 0.32, 0.68} &
\single{0.44} \\
\midrule
\multicolumn{9}{|c|}{\textbf{Open-source Models}} \\ \hline
\multicolumn{9}{|c|}{\textbf{Huge Models}} \\ 
\midrule
DeepSeek-V3 &
\scorecell{(0.79+0.78+0.78)/3}{0.79, 0.78, 0.78} &
\scorecell{(0.97+0.96+0.94)/3}{0.97, 0.96, 0.94} &
\scorecell{(1.00+1.00+0.00)/3}{1.00, 1.00, 0.00} &
\single{0.78} &
\scorecell{(0.76+0.75+0.75)/3}{0.76, 0.75, 0.75} &
\scorecell{(0.77+0.73+0.75)/3}{0.77, 0.73, 0.75} &
\scorecell{(0.81+0.83+0.17)/3}{0.81, 0.83, 0.17} &
\single{0.82} \\

DeepSeek-R1 &
\scorecell{(0.70+0.65+0.67)/3}{0.70, 0.65, 0.67} &
\scorecell{(0.00+0.00+0.00)/3}{0.00, 0.00, 0.00} &
\scorecell{(0.02+0.08+0.92)/3}{0.02, 0.08, 0.92} &
\single{0.03} &
\scorecell{(0.68+0.63+0.65)/3}{0.68, 0.63, 0.65} &
\scorecell{(0.00+0.00+0.00)/3}{0.00, 0.00, 0.00} &
\scorecell{(0.01+0.01+0.97)/3}{0.01, 0.01, 0.97} &
\single{0.00} \\

Qwen3 235B &
\scorecell{(0.62+0.68+0.65)/3}{0.62, 0.68, 0.65} &
\scorecell{(0.01+0.00+0.00)/3}{0.01, 0.00, 0.00} &
\scorecell{(0.02+0.09+0.91)/3}{0.02, 0.09, 0.91} &
\single{0.00} &
\scorecell{(0.64+0.69+0.66)/3}{0.64, 0.69, 0.66} &
\scorecell{(0.08+0.00+0.00)/3}{0.08, 0.00, 0.00} &
\scorecell{(0.04+0.08+0.92)/3}{0.04, 0.08, 0.92} &
\single{0.01} \\

LLaMA-4 Maverick &
\scorecell{(0.65+0.69+0.67)/3}{0.65, 0.69, 0.67} &
\scorecell{(0.98+0.90+0.97)/3}{0.98, 0.90, 0.97} &
\scorecell{(0.57+0.66+0.34)/3}{0.57, 0.66, 0.34} &
\single{0.37} &
\scorecell{(0.73+0.64+0.68)/3}{0.73, 0.64, 0.68} &
\scorecell{(0.98+0.95+0.94)/3}{0.98, 0.95, 0.94} &
\scorecell{(0.65+0.68+0.32)/3}{0.65, 0.68, 0.32} &
\single{0.62} \\

LLaMA-4 Scout &
\scorecell{(0.75+0.77+0.76)/3}{0.75, 0.77, 0.76} &
\scorecell{(1.00+0.93+0.95)/3}{1.00, 0.93, 0.95} &
\scorecell{(0.66+0.73+0.27)/3}{0.66, 0.73, 0.27} &
\single{0.41} &
\scorecell{(0.79+0.80+0.79)/3}{0.79, 0.80, 0.79} &
\scorecell{(1.00+0.95+0.97)/3}{1.00, 0.95, 0.97} &
\scorecell{(0.56+0.64+0.36)/3}{0.56, 0.64, 0.36} &
\single{0.44} \\

\midrule
\multicolumn{9}{|c|}{\textbf{Large Models}} \\ 
\midrule
DeepSeek-R1-70B &
\scorecell{(0.71+0.71+0.71)/3}{0.71, 0.71, 0.71} &
\scorecell{(0.00+0.00+0.00)/3}{0.00, 0.00, 0.00} &
\scorecell{(0.02+0.03+0.95)/3}{0.02, 0.03, 0.95} &
\single{0.19} &
\scorecell{(0.64+0.72+0.68)/3}{0.64, 0.72, 0.68} &
\scorecell{(0.00+0.00+0.00)/3}{0.00, 0.00, 0.00} &
\scorecell{(0.03+0.09+0.91)/3}{0.03, 0.09, 0.91} &
\single{0.00} \\

LLaMA-3-70B &
\scorecell{(0.72+0.65+0.68)/3}{0.72, 0.65, 0.68} &
\scorecell{(0.17+0.11+0.12)/3}{0.17, 0.11, 0.12} &
\scorecell{(0.17+0.20+0.80)/3}{0.17, 0.20, 0.80} &
\single{0.43} &
\scorecell{(0.70+0.66+0.68)/3}{0.70, 0.66, 0.68} &
\scorecell{(0.41+0.39+0.39)/3}{0.41, 0.39, 0.39} &
\scorecell{(0.48+0.55+0.45)/3}{0.48, 0.55, 0.45} &
\single{0.20} \\

\midrule
\multicolumn{9}{|c|}{\textbf{Medium Models}} \\
\midrule

Qwen QwQ 32B &
\scorecell{(0.94+0.92+0.93)/3}{0.94, 0.92, 0.93} &
\scorecell{(1.00+0.96+1.00)/3}{1.00, 0.96, 1.00} &
\scorecell{(0.87+0.89+0.11)/3}{0.87, 0.89, 0.11} &
\single{0.84} &
\scorecell{(0.86+0.93+0.89)/3}{0.86, 0.93, 0.89} &
\scorecell{(1.00+0.97+1.00)/3}{1.00, 0.97, 1.00} &
\scorecell{(0.57+0.65+0.35)/3}{0.57, 0.65, 0.35} &
\single{0.64} \\

Qwen3-30B &
\scorecell{(0.74+0.68+0.71)/3}{0.74, 0.68, 0.71} &
\scorecell{(0.04+0.04+0.03)/3}{0.04, 0.04, 0.03} &
\scorecell{(0.05+0.06+0.94)/3}{0.05, 0.06, 0.94} &
\single{0.19} &
\scorecell{(0.74+0.62+0.67)/3}{0.74, 0.62, 0.67} &
\scorecell{(0.00+0.00+0.00)/3}{0.00, 0.00, 0.00} &
\scorecell{(0.01+0.04+0.96)/3}{0.01, 0.04, 0.96} &
\single{0.20} \\

Gemma3-27B &
\scorecell{(0.30+0.38+0.34)/3}{0.30, 0.38, 0.34} &
\scorecell{(0.00+0.00+0.00)/3}{0.00, 0.00, 0.00} &
\scorecell{(0.00+0.03+0.97)/3}{0.00, 0.03, 0.97} &
\single{0.01} &
\scorecell{(0.26+0.34+0.29)/3}{0.26, 0.34, 0.29} &
\scorecell{(0.00+0.00+0.00)/3}{0.00, 0.00, 0.00} &
\scorecell{(0.00+0.02+0.95)/3}{0.00, 0.02, 0.95} &
\single{0.04} \\

\midrule
\multicolumn{9}{|c|}{\textbf{Small Models}} \\
\midrule

Gemma-2-9B &
\scorecell{(0.69+0.67+0.68)/3}{0.69, 0.67, 0.68} &
\scorecell{(0.24+0.15+0.19)/3}{0.24, 0.15, 0.19} &
\scorecell{(0.19+0.23+0.77)/3}{0.19, 0.23, 0.77} &
\single{0.24} &
\scorecell{(0.60+0.72+0.65)/3}{0.60, 0.72, 0.65} &
\scorecell{(0.24+0.15+0.17)/3}{0.24, 0.15, 0.17} &
\scorecell{(0.17+0.21+0.79)/3}{0.17, 0.21, 0.79} &
\single{0.19} \\

LLaMA-3-8B &
\scorecell{(0.72+0.65+0.68)/3}{0.72, 0.65, 0.68} &
\scorecell{(1.00+0.92+0.98)/3}{1.00, 0.92, 0.98} &
\scorecell{(0.38+0.44+0.56)/3}{0.38, 0.44, 0.56} &
\single{0.20} &
\scorecell{(0.71+0.61+0.66)/3}{0.71, 0.61, 0.66} &
\scorecell{(0.98+0.95+0.97)/3}{0.98, 0.95, 0.97} &
\scorecell{(0.45+0.51+0.49)/3}{0.45, 0.51, 0.49} &
\single{0.19} \\

Qwen-3-14B &
\scorecell{(0.70+0.69+0.69)/3}{0.70, 0.69, 0.69} &
\scorecell{(0.04+0.00+0.04)/3}{0.04, 0.00, 0.04} &
\scorecell{(0.06+0.11+0.89)/3}{0.06, 0.11, 0.89} &
\single{0.02} &
\scorecell{(0.59+0.65+0.62)/3}{0.59, 0.65, 0.62} &
\scorecell{(0.00+0.00+0.00)/3}{0.00, 0.00, 0.00} &
\scorecell{(0.03+0.07+0.93)/3}{0.03, 0.07, 0.93} &
\single{0.04} \\

Gemma3-12B-instruct &
\scorecell{(0.38+0.36+0.37)/3}{0.38, 0.36, 0.37} &
\scorecell{(0.00+0.00+0.00)/3}{0.00, 0.00, 0.00} &
\scorecell{(0.00+0.05+0.95)/3}{0.00, 0.05, 0.95} &
\single{0.05} &
\scorecell{(0.34+0.37+0.35)/3}{0.34, 0.37, 0.35} &
\scorecell{(0.00+0.00+0.00)/3}{0.00, 0.00, 0.00} &
\scorecell{(0.06+0.08+0.92)/3}{0.06, 0.08, 0.92} &
\single{0.04} \\

\bottomrule
\end{tabular}}
\end{table*}

\section{Experiments and Results}
\subsection{Implementation and Evaluation Details}
We utilize the LangGraph architecture \cite{Langgraph} for agent construction and workflow graph compilation. Each agent is assigned a tailored toolset, drawn from our proposed suite of 16 tools \textbf{(see Appendix B.1)}, with the selection guided by the agent’s specific role and the need to omit redundant or irrelevant functionalities. In order to assess the capabilities of existing LLM agents, we validate a total number of 24 models on the FedAgentBench datasets, including:
(1) 10 representative proprietary LLMs: GPT 4.1, GPT-4o, GPT-4, GPT-4-Turbo, GPT 4.1-mini, GPT-4o-mini, GPT o4-mini, GPT o3-mini, GPT-3.5 Turbo, and Claude-3.7 Sonnet. (2) 14 state-of-the-art open-sourced LLMs ranging from 9B to 685B: LLaMA series models (LLaMA-4 Maverick, LLaMA-4 Scout, LLaMA-3 70B, LLaMA-3 8B), DeepSeek series models (DeepSeek-V3, Deepseek-R1, DeepSeek-R1-Distill-Llama-70B), Qwen series models (Qwen 3 235B, Qwen QwQ 32B, Qwen 3 30B, Qwen 3 14B) and Gemma series models (Gemma 3 27B Instruct, Gemma 3 12B Instruct, Gemma 2 9B Instruct). We utilize APIs from \cite{openai_api_2025}, \cite{groq_models_2025}, \cite{deepinfra_models_2025}.

\textbf{Evaluation metrics:} We evaluate the agentic performance using a total of \textbf{13 key metrics} in different steps of the FL workflow: \textbf{(1) For each step}, we use \textbf{Success Rate over 5 runs} which is a binary indicator of task success/completion. It evaluates the ability of the multi-agent framework to generate executable outputs that satisfy the task requirements. \textbf{(2) For client selection step,} we use \textbf{Precision, Recall, and F1 score} of selected clients vs. the canonical eligible client set. \textbf{(3) For data pre-processing step}, we use \textbf{(i) Schema Compliance Rate}, \textit{i.e.}, proportion of correctly structured folders/files, \textbf{(ii) Duplicate Removal Rate}, \textit{i.e.}, proportion of duplicates removed, and \textbf{(iii) Format Normalization Rate}, \textit{i.e.},  proportion of files correctly normalized (e.g., format, resolution). \textbf{(4) For label harmonization step}, we use: \textbf{(i) Exact-match Accuracy} of label mappings vs. the canonical schema, \textbf{(ii) Coverage Rate}, \textit{i.e.},  proportion of local classes successfully mapped, \textbf{(iii) Conflict Rate}, \textit{i.e.}, proportion of classes with ambiguous mappings. \textbf{(5) For federated training step}, we use \textbf{Training Start Verification} as the metric to determine whether the agent produces valid configuration files, initializes the training process, and logs the start signal. Besides, \textbf{for each step}, we also compute \textbf{(6) Time Spent in seconds} which denotes the duration required to complete the task \textbf{(see Appendix D \& Table 16 for comparison of average time)}; and \textbf{(7) Token Requirement} which indicates the number of tokens involved \textbf{(see Fig. 1 (d) for comparison of token requirement)}.



\subsection{Main Results and Key Insights}
\begin{wrapfigure}{r}{.85\textwidth}
  \centering 
  \includegraphics[width=\linewidth]{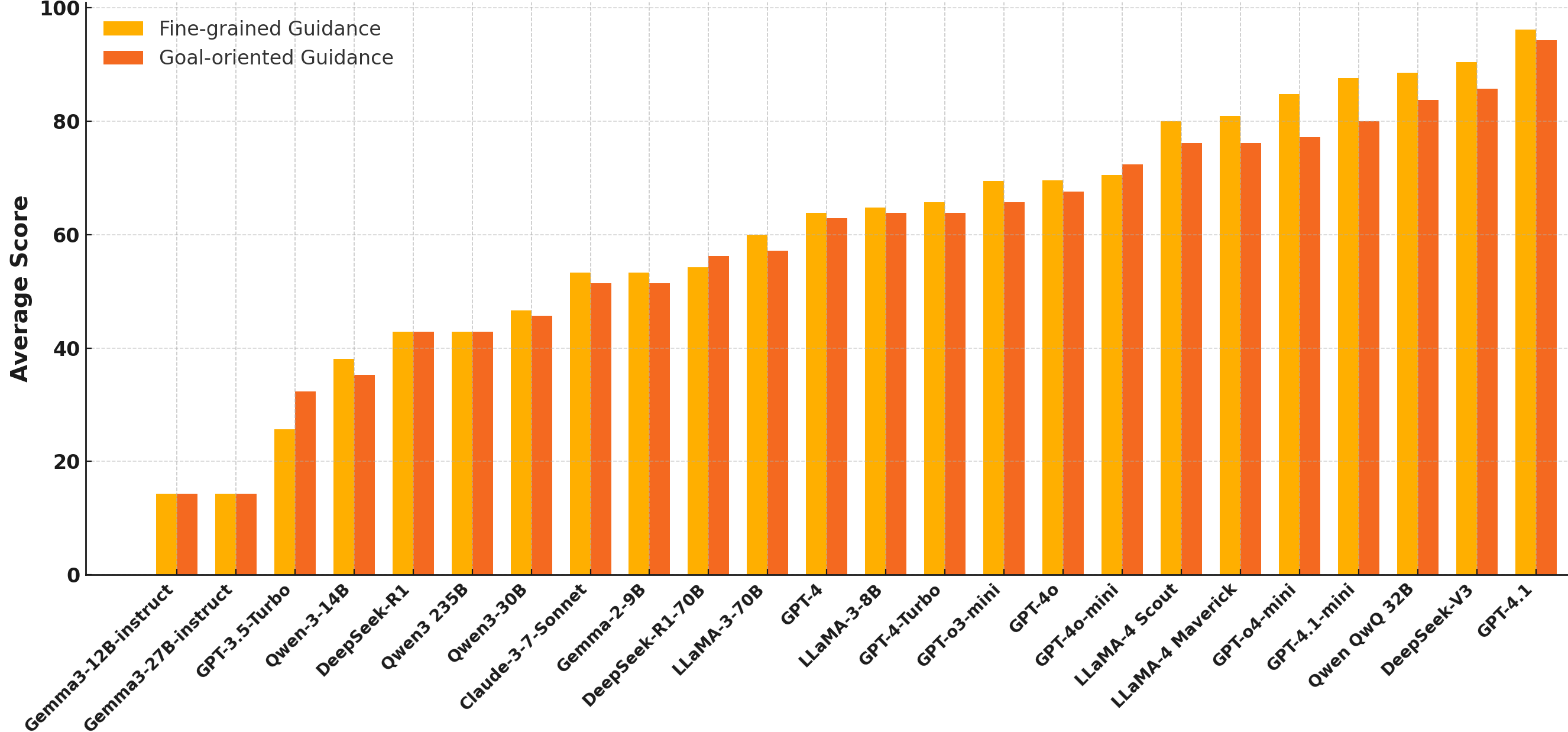} 
    \caption{Overall performance of LLM agents in FedAgentBench}
    \label{fig:prior_exp}
\end{wrapfigure}
We summarize the overall success scores of all agent cores over 6 modality specific environments with two types of guidance styles for prompting LLMs \textit{viz.}, fine-grained guidance (explicit step-by-step instructions) and goal-oriented guidance (high-level task description) in Fig. 7. We also show detailed performance breakdown of Dermatology environment in Table 1 and Histopathology in Table 2.\textbf{ For detailed results in all other environments, please see Appendix D. \& Tables 10-15} Also, see Fig 1 (d) for overall token requirements of each model.

From the tables, we find proprietary models consistently outperform open-source ones across all FL stages. Besides, fine-grained guidance yields higher success rates than goal-oriented prompts for most models. Performance drops in more complex tasks like label harmonization compared to client selection. We also observe that model size alone does not guarantee performance (see Fig. 7). Instead, architectural design and instruction-following capability are more critical.

\textbf{Proprietary Model Performance:} GPT-4.1 and GPT-4.1-mini show top-tier performance (85–100\%), especially under fine-grained guidance. GPT-4o, although newer, struggles with label harmonization and federated training across all environments, leading to overall lower scores (~62-71\%). Claude-3.7-Sonnet achieves moderate performance (51–57\%), inferior to GPT-4 variants. GPT-3.5-Turbo and older variants perform poorly, barely completing the complex stages.

\textbf{Open-source Model Performance:} We discuss agent performance based on model sizes below:\\
\textbf{(i) Huge Models:} DeepSeek-V3 is the strongest open-source model contender with 80–94\% success rate comparable to the best proprietary models. Qwen3 and DeepSeek-R1 perform inconsistently, often failing in more structured stages like data pre-processing and label harmonization. \\
\textbf{(ii) Medium and Large Models:} Qwen QwQ 32B demonstrates strong performance (82–91\%) and outperforms several proprietary models even under goal-oriented setups. LLaMA-4 Scout and Maverick also deliver competitive performance, especially in label harmonization and federated training, with scores in the 71–94\% range
Other large models such as LLaMA-3-70B, and Qwen3-30B struggle with most tasks except initial client communication or final training step. Gemma3-27B-instruct is unusable under almost all these settings. \\
\textbf{(iii) Small Models:} Performance of 8-14B sized-models drops significantly. Most models (except LLaMA 3 8B) achieve less than or around 50\% success. Particularly, Gemma 3-12B-instruct and Qwen 3 14B are observed to fail due to extreme hallucinations. These models are unable to perform any label-oriented reasoning and structured data operations, even under fine-grained instructions.

\textbf{Impact of Task Complexity:} High success is observed in the initial and final steps of client orchestration and federated training across almost all agents, including weaker ones indicating that these tasks are relatively simpler. Data Pre-processing and Label Harmonization are seen to be major differentiators among agents. Weaker agents particularly fail to perform these tasks especially in goal-oriented scenarios, where planning and file structure comprehension are needed. Across almost all agents, label harmonization shows lowest success rates, regardless of guidance type. This suggests that aligning semantic labels across clients remains one of the hardest challenges. Among modalities, histopathology has the highest semantic complexity, potentially due to domain-specific terminology.

\textbf{Granularity of guidance:} In fine-grained guidance, we provide explicit instruction to the models to follow a particular workflow whereas in goal-oriented guidance, we mention the overall objective of the agent without specifying the exact steps, thereby requiring autonomous planning or reasoning. Fine-grained guidance is seen to outperform goal-oriented guidance across almost every model, especially for weaker agents. More capable models like GPT-4.1 and DeepSeek-V3 close this gap, showing their capability to plan even based on implicit prompts.

\begin{table*}[t]
\centering
\scriptsize
\caption{Comparison in terms of success rate over 5 runs in \textbf{Histopathology} environment}

\begingroup
\setlength{\tabcolsep}{3pt}
\renewcommand{\arraystretch}{0.80}
\addtolength{\aboverulesep}{-0.3ex}
\addtolength{\belowrulesep}{-0.3ex}
\begin{tabular}{|@{}l@{}|@{}c@{}c@{}c@{}c@{}|c|@{}c@{}c@{}c@{}c@{}|c|}

\hline
\textbf{Model} & \multicolumn{5}{c|}{\textbf{Fine-grained guidance}} & \multicolumn{5}{c|}{\textbf{Goal-oriented guidance}} \\
\hline
  & \textbf{Client-Sel} & \textbf{Data-Pre} & \textbf{Label-Harm} & \textbf{Fed-Train} & \textbf{Overall} & \textbf{Client-Sel} & \textbf{Data-Pre} & \textbf{Label-Harm} & \textbf{Fed-Train} & \textbf{Overall} 
  \\ \hline
      & \textbf{$S_1$, $C_1$, $S_2$} & \textbf{$C_2$} & \textbf{$C_3$} & \textbf{$S_3$, $S_4$} & \textbf{} & \textbf{$S_1$, $C_1$, $S_2$} & \textbf{$C_2$} & \textbf{$C_3$} & \textbf{$S_3$, $S_4$} & \textbf{} \\
\hline
\multicolumn{11}{|c|}{\textbf{Proprietary Models}} \\ \hline
GPT-4.1 &
\chip{5}{5}, \chip{4}{5}, \chip{5}{5} & \chip{5}{5} & \chip{5}{5} & \chip{4}{5}, \chip{5}{5} & \scorecellpct{94.29} &
\chip{5}{5}, \chip{4}{5}, \chip{5}{5} & \chip{5}{5} & \chip{5}{5} & \chip{4}{5}, \chip{5}{5} & \scorecellpct{94.29} \\

GPT-4o &
\chip{5}{5}, \chip{0}{5}, \chip{5}{5} & \chip{5}{5} & \chip{2}{5} & \chip{1}{5}, \chip{5}{5} & \scorecellpct{65.71} &
\chip{5}{5}, \chip{0}{5}, \chip{5}{5} & \chip{5}{5} & \chip{1}{5} & \chip{1}{5}, \chip{5}{5} & \scorecellpct{62.86} \\

GPT-4 &
\chip{5}{5}, \chip{1}{5}, \chip{5}{5} & \chip{0}{5} & \chip{1}{5} & \chip{2}{5}, \chip{5}{5} & \scorecellpct{54.29} &
\chip{5}{5}, \chip{1}{5}, \chip{5}{5} & \chip{0}{5} & \chip{0}{5} & \chip{2}{5}, \chip{5}{5} & \scorecellpct{51.43} \\

GPT-4-Turbo &
\chip{5}{5}, \chip{1}{5}, \chip{5}{5} & \chip{1}{5} & \chip{1}{5} & \chip{2}{5}, \chip{5}{5} & \scorecellpct{57.14} &
\chip{5}{5}, \chip{1}{5}, \chip{5}{5} & \chip{4}{5} & \chip{1}{5} & \chip{2}{5}, \chip{5}{5} & \scorecellpct{65.71} \\

GPT-4.1-mini &
\chip{5}{5}, \chip{3}{5}, \chip{5}{5} & \chip{5}{5} & \chip{4}{5} & \chip{3}{5}, \chip{5}{5} & \scorecellpct{85.71} &
\chip{5}{5}, \chip{3}{5}, \chip{5}{5} & \chip{3}{5} & \chip{4}{5} & \chip{3}{5}, \chip{5}{5} & \scorecellpct{80.00} \\

GPT-4o-mini &
\chip{5}{5}, \chip{1}{5}, \chip{3}{5} & \chip{5}{5} & \chip{3}{5} & \chip{2}{5}, \chip{4}{5} & \scorecellpct{65.71} &
\chip{5}{5}, \chip{1}{5}, \chip{3}{5} & \chip{5}{5} & \chip{1}{5} & \chip{2}{5}, \chip{4}{5} & \scorecellpct{60.00} \\

GPT-o4-mini &
\chip{5}{5}, \chip{2}{5}, \chip{5}{5} & \chip{5}{5} & \chip{3}{5} & \chip{2}{5}, \chip{5}{5} & \scorecellpct{77.14} &
\chip{5}{5}, \chip{2}{5}, \chip{5}{5} & \chip{4}{5} & \chip{2}{5} & \chip{2}{5}, \chip{4}{5} & \scorecellpct{68.57} \\

GPT-o3-mini &
\chip{5}{5}, \chip{5}{5}, \chip{5}{5} & \chip{0}{5} & \chip{2}{5} & \chip{3}{5}, \chip{5}{5} & \scorecellpct{71.43} &
\chip{5}{5}, \chip{4}{5}, \chip{5}{5} & \chip{0}{5} & \chip{2}{5} & \chip{3}{5}, \chip{5}{5} & \scorecellpct{68.57} \\

GPT-3.5-Turbo &
\chip{5}{5}, \chip{0}{5}, \chip{0}{5} & \chip{0}{5} & \chip{0}{5} & \chip{1}{5}, \chip{3}{5} & \scorecellpct{25.71} &
\chip{5}{5}, \chip{0}{5}, \chip{0}{5} & \chip{2}{5} & \chip{0}{5} & \chip{1}{5}, \chip{3}{5} & \scorecellpct{31.43} \\

Claude-3-7-Sonnet &
\chip{5}{5}, \chip{2}{5}, \chip{3}{5} & \chip{2}{5} & \chip{1}{5} & \chip{2}{5}, \chip{3}{5} & \scorecellpct{51.43} &
\chip{5}{5}, \chip{2}{5}, \chip{3}{5} & \chip{2}{5} & \chip{1}{5} & \chip{2}{5}, \chip{5}{5} & \scorecellpct{57.14} \\

\hline
\multicolumn{11}{|c|}{\textbf{Open-source Models}} \\ \hline
\multicolumn{11}{|c|}{\textbf{Huge Models}} \\ \hline
DeepSeek-V3 &
\chip{5}{5}, \chip{3}{5}, \chip{5}{5} & \chip{5}{5} & \chip{5}{5} & \chip{4}{5}, \chip{5}{5} & \scorecellpct{91.43} &
\chip{5}{5}, \chip{3}{5}, \chip{5}{5} & \chip{4}{5} & \chip{5}{5} & \chip{4}{5}, \chip{5}{5} & \scorecellpct{88.57} \\
DeepSeek-R1 &
\chip{5}{5}, \chip{0}{5}, \chip{5}{5} & \chip{0}{5} & \chip{0}{5} & \chip{0}{5}, \chip{5}{5} & \scorecellpct{42.86} &
\chip{5}{5}, \chip{0}{5}, \chip{5}{5} & \chip{0}{5} & \chip{0}{5} & \chip{0}{5}, \chip{5}{5} & \scorecellpct{42.86} \\
Qwen3 235B &
\chip{5}{5}, \chip{0}{5}, \chip{5}{5} & \chip{0}{5} & \chip{0}{5} & \chip{0}{5}, \chip{5}{5} & \scorecellpct{42.86} &
\chip{5}{5}, \chip{0}{5}, \chip{5}{5} & \chip{0}{5} & \chip{0}{5} & \chip{0}{5}, \chip{5}{5} & \scorecellpct{42.86} \\
LLaMA-4 Maverick &
\chip{5}{5}, \chip{2}{5}, \chip{4}{5} & \chip{5}{5} & \chip{3}{5} & \chip{3}{5}, \chip{5}{5} & \scorecellpct{77.14} &
\chip{5}{5}, \chip{2}{5}, \chip{4}{5} & \chip{5}{5} & \chip{3}{5} & \chip{3}{5}, \chip{5}{5} & \scorecellpct{71.43} \\
LLaMA-4 Scout &
\chip{5}{5}, \chip{2}{5}, \chip{5}{5} & \chip{5}{5} & \chip{4}{5} & \chip{2}{5}, \chip{5}{5} & \scorecellpct{80.00} &
\chip{5}{5}, \chip{2}{5}, \chip{5}{5} & \chip{5}{5} & \chip{3}{5} & \chip{2}{5}, \chip{5}{5} & \scorecellpct{77.14} \\ \hline

\multicolumn{11}{|c|}{\textbf{Large Models}} \\ \hline
DeepSeek-R1-70B &
\chip{5}{5}, \chip{0}{5}, \chip{5}{5} & \chip{0}{5} & \chip{0}{5} & \chip{0}{5}, \chip{5}{5} & \scorecellpct{42.86} &
\chip{5}{5}, \chip{0}{5}, \chip{5}{5} & \chip{0}{5} & \chip{0}{5} & \chip{0}{5}, \chip{5}{5} & \scorecellpct{42.86} \\
LLaMA-3-70B &
\chip{5}{5}, \chip{1}{5}, \chip{5}{5} & \chip{1}{5} & \chip{1}{5} & \chip{1}{5}, \chip{5}{5} & \scorecellpct{54.29} &
\chip{5}{5}, \chip{1}{5}, \chip{5}{5} & \chip{2}{5} & \chip{2}{5} & \chip{1}{5}, \chip{5}{5} & \scorecellpct{60.00} \\ \hline

\multicolumn{11}{|c|}{\textbf{Medium Models}} \\ \hline
Qwen QwQ 32B &
\chip{5}{5}, \chip{4}{5}, \chip{5}{5} & \chip{3}{5} & \chip{4}{5} & \chip{4}{5}, \chip{5}{5} & \scorecellpct{85.71} &
\chip{5}{5}, \chip{4}{5}, \chip{5}{5} & \chip{2}{5} & \chip{4}{5} & \chip{4}{5}, \chip{5}{5} & \scorecellpct{82.86} \\
Qwen3-30B &
\chip{5}{5}, \chip{0}{5}, \chip{5}{5} & \chip{0}{5} & \chip{0}{5} & \chip{1}{5}, \chip{5}{5} & \scorecellpct{45.71} &
\chip{5}{5}, \chip{0}{5}, \chip{5}{5} & \chip{0}{5} & \chip{0}{5} & \chip{1}{5}, \chip{5}{5} & \scorecellpct{45.71} \\
Gemma3-27B-instruct &
\chip{5}{5}, \chip{0}{5}, \chip{0}{5} & \chip{0}{5} & \chip{0}{5} & \chip{0}{5}, \chip{0}{5} & \scorecellpct{14.29} &
\chip{5}{5}, \chip{0}{5}, \chip{0}{5} & \chip{0}{5} & \chip{0}{5} & \chip{0}{5}, \chip{0}{5} & \scorecellpct{14.29} \\ \hline

\multicolumn{11}{|c|}{\textbf{Small Models}} \\ \hline
Gemma-2-9B &
\chip{5}{5}, \chip{1}{5}, \chip{5}{5} & \chip{2}{5} & \chip{1}{5} & \chip{1}{5}, \chip{5}{5} & \scorecellpct{57.14} &
\chip{5}{5}, \chip{1}{5}, \chip{5}{5} & \chip{1}{5} & \chip{1}{5} & \chip{1}{5}, \chip{5}{5} & \scorecellpct{54.29} \\
LLaMA-3-8B &
\chip{5}{5}, \chip{0}{5}, \chip{5}{5} & \chip{5}{5} & \chip{2}{5} & \chip{1}{5}, \chip{5}{5} & \scorecellpct{65.71} &
\chip{5}{5}, \chip{0}{5}, \chip{5}{5} & \chip{5}{5} & \chip{2}{5} & \chip{1}{5}, \chip{5}{5} & \scorecellpct{65.71} \\
Qwen-3-14B &
\chip{5}{5}, \chip{0}{5}, \chip{5}{5} & \chip{0}{5} & \chip{0}{5} & \chip{0}{5}, \chip{5}{5} & \scorecellpct{42.86} &
\chip{5}{5}, \chip{0}{5}, \chip{5}{5} & \chip{0}{5} & \chip{0}{5} & \chip{0}{5}, \chip{4}{5} & \scorecellpct{40.00} \\
Gemma3-12B-instruct &
\chip{5}{5}, \chip{0}{5}, \chip{0}{5} & \chip{0}{5} & \chip{0}{5} & \chip{0}{5}, \chip{0}{5} & \scorecellpct{14.29} &
\chip{5}{5}, \chip{0}{5}, \chip{0}{5} & \chip{0}{5} & \chip{0}{5} & \chip{0}{5}, \chip{0}{5} & \scorecellpct{14.29} \\ 

\hline
\end{tabular}
\label{tab:llm-fl-instruction}
\endgroup
\end{table*}

\subsection{Agent Failure Analysis:}
We identify six key recurring failure modes of LLM agents across FL sub-tasks that highlight important limitations of current LLM capabilities in FL workflows \textbf{(see Appendix D for more details)}: \\
\textbf{(i) Lack of Domain-Specific Reasoning:} The agents frequently fail to apply relevant medical domain knowledge.\textbf{ Eg:} In label harmonization (Fig 6), the agents often miss subtle mismatches between dermatology folder names and coarse class labels possibly due to the lack of domain grounding and inability to handle naming conventions specific to medical datasets.\\
\textbf{(ii) Failure in Multi-Step Planning:} The agents are often unable to follow multi-step workflows, skipping essential operations where multiple sequential actions are required.\textbf{ Eg:} Data pre-processor agents often overlook file/data cleaning steps of Fig. 5 due to multiple tasks in single execution cycle.\\
\textbf{(iii) Overconfidence and Shortcutting:} The agents recurrently provide wrong solutions, by defaulting to plausible but incorrect logic when unsure, instead of expressing uncertainty. \textbf{Eg:} Assigning both “nevus” and “melanoma metastasis” to the 'benign' class to simplify label mapping.\\
\textbf{(iv) Hallucination in Structured Multi-Agent Tasks:} The agents (particularly DeepSeek R1 and Gemma-based models) often generate irrelevant or unrelated outputs despite specific instructions due to misalignment with structured task formats and poor control over output scope \textbf{(see Fig. 17-18 in Appendix D)}. \textbf{Eg:} When asked to select skin cancer dataset, Gemma-3 27B Instruct repeatedly returned philosophical or sarcastic monologues in foreign languages, tutorials on freelancing, etc.\\
\textbf{(v) Task-Type and Modality Mismatch Due to Prior Assumptions:} Agents can sometimes confuse tasks or ignore modality constraints due to frequency biases and shallow keyword matching instead of hierarchical task understanding. \textbf{Eg:} Recommending a malignant lesion segmentation dataset for a classification task or ultrasound datasets for histopathology-based breast cancer detection task.\\
\textbf{(vi) Procedural Overthinking and Paralysis by Analysis:} The reasoning/thinking agents often delay execution by speculating about dataset structure or missing dependencies without being asked, potentially due to excessive internal reasoning without grounding in file system or available information \textbf{(see Fig. 15 in Appendix D)}. \textbf{Eg:} DeepSeek R1 repeatedly debates whether a client dataset should be selected without reading the dataset description file.
\section{Conclusion and Limitation}
In this paper, we introduced \textbf{the first agent-driven FL framework} and an associated benchmark, \textbf{FedAgentBench}, for evaluating LLM agents across diverse tasks constituting typical FL workflows. The evaluation covers 24 LLMs with varying sizes and a wide range of FL sub-tasks with varying difficulty levels in six modality-specific FL settings that closely simulate real-world clinical FL environments. Our framework is privacy preserving, comprehensive and modular. It includes 201 medical datasets and 40 FL algorithms and can be easily extended to incorporate more functionalities, settings, and algorithms specific to the user requirement.

We investigated the impact of various factors like FL task complexity and granularity of guidance on the agent performance and analyzed the common failure modes of different agents. Our experiments reveal that across all environments, GPT-4.1 achieves almost perfect scores, under both fine-grained and goal-oriented prompting, whereas GPT-3.5-Turbo, Gemma3 series, and some Qwen variants consistently underperform across all stages and environments. DeepSeek-V3, Qwen QwQ 32B, and LLaMA-4 Maverick are the most reliable open-source agents across tasks. Unsurprisingly, fine-grained guidance consistently outperforms goal-oriented prompting, especially for less capable models. Our findings highlight that the order of complexity of the FL sub-tasks for most agents is: Label Harmonization > Data Pre-processing > Federated Training > Client Orchestration. Our experiments also show that larger model size does not necessarily correlate with better performance, \textit{i.e.}, some mid-sized models (30–40B) outperform larger ones (70B+). E.g., Qwen QwQ 32B consistently outperforms Qwen3-235B and DeepSeek-R1-70B.




This work is a first step toward agent-driven FL and comes with some limitations: (a) We currently assume stable network conditions and do not model dynamic communication bandwidth. (b) We do not incorporate real-time monitoring or interruption mechanisms. (c) We do not simulate safety check or regulatory compliance assessment but it can be seamlessly integrated into the system.

\section{Acknowledgments}
This work was supported in part by the UK EPSRC (Engineering and Physical Research Council) Programme Grant EP/T028572/1 (VisualAI), a  UK EPSRC Doctoral Training Partnership award, the UKRI grant EP/X040186/1 (Turing AI Fellowship), and the InnoHK-funded Hong Kong Centre for Cerebro-cardiovascular Health Engineering (COCHE) Project 2.1 (Cardiovascular risks in early life and fetal echocardiography).

\bibliography{ref}

\begin{thebibliography}{17}
\providecommand{\natexlab}[1]{#1}
\providecommand{\url}[1]{\texttt{#1}}
\expandafter\ifx\csname urlstyle\endcsname\relax
  \providecommand{\doi}[1]{doi: #1}\else
  \providecommand{\doi}{doi: \begingroup \urlstyle{rm}\Url}\fi

\bibitem[Cai et~al.()Cai, Wang, Ma, Chen, and Zhou]{cailarge}
Tianle Cai, Xuezhi Wang, Tengyu Ma, Xinyun Chen, and Denny Zhou.
\newblock Large language models as tool makers.
\newblock In \emph{The Twelfth International Conference on Learning Representations}.

\bibitem[Chu et~al.(2025)Chu, Wang, Xie, Zhu, Yan, Ye, Zhong, Hu, Liang, Yu, et~al.]{chu2025llm}
Zhendong Chu, Shen Wang, Jian Xie, Tinghui Zhu, Yibo Yan, Jinheng Ye, Aoxiao Zhong, Xuming Hu, Jing Liang, Philip~S Yu, et~al.
\newblock Llm agents for education: Advances and applications.
\newblock \emph{arXiv preprint arXiv:2503.11733}, 2025.

\bibitem[{Deep Infra}(2025)]{deepinfra_models_2025}
{Deep Infra}.
\newblock Deepinfra models documentation, 2025.
\newblock URL \url{https://deepinfra.com/docs/models}.
\newblock Accessed: 2025-05-16.

\bibitem[Gou et~al.()Gou, Shao, Gong, Yang, Huang, Duan, Chen, et~al.]{goutora}
Zhibin Gou, Zhihong Shao, Yeyun Gong, Yujiu Yang, Minlie Huang, Nan Duan, Weizhu Chen, et~al.
\newblock Tora: A tool-integrated reasoning agent for mathematical problem solving.
\newblock In \emph{The Twelfth International Conference on Learning Representations}.

\bibitem[{Groq}(2025)]{groq_models_2025}
{Groq}.
\newblock Groqcloud supported models, 2025.
\newblock URL \url{https://console.groq.com/docs/models}.
\newblock Accessed: 2025-05-16.

\bibitem[Gur et~al.()Gur, Furuta, Huang, Safdari, Matsuo, Eck, and Faust]{gurreal}
Izzeddin Gur, Hiroki Furuta, Austin~V Huang, Mustafa Safdari, Yutaka Matsuo, Douglas Eck, and Aleksandra Faust.
\newblock A real-world webagent with planning, long context understanding, and program synthesis.
\newblock In \emph{The Twelfth International Conference on Learning Representations}.

\bibitem[{Langgraph}(2025)]{Langgraph}
{Langgraph}.
\newblock Langgraph documentation, 2025.
\newblock URL \url{https://www.langchain.com/langgraph}.
\newblock Accessed: 2025-05-16.

\bibitem[Li et~al.(2023)Li, Hammoud, Itani, Khizbullin, and Ghanem]{li2023camel}
Guohao Li, Hasan Hammoud, Hani Itani, Dmitrii Khizbullin, and Bernard Ghanem.
\newblock Camel: Communicative agents for" mind" exploration of large language model society.
\newblock \emph{Advances in Neural Information Processing Systems}, 36:\penalty0 51991--52008, 2023.

\bibitem[Li et~al.(2021)Li, Wen, Wu, Hu, Wang, Li, Liu, and He]{li2021survey}
Qinbin Li, Zeyi Wen, Zhaomin Wu, Sixu Hu, Naibo Wang, Yuan Li, Xu~Liu, and Bingsheng He.
\newblock A survey on federated learning systems: Vision, hype and reality for data privacy and protection.
\newblock \emph{IEEE Transactions on Knowledge and Data Engineering}, pp.\  1--20, 2021.

\bibitem[Li et~al.(2020)Li, Sahu, Talwalkar, and Smith]{li2020federated}
Tian Li, Anit~Kumar Sahu, Ameet Talwalkar, and Virginia Smith.
\newblock Federated optimization in heterogeneous networks.
\newblock In \emph{Proceedings of Machine Learning and Systems (MLSys)}, 2020.

\bibitem[Luo et~al.(2025)Luo, Zhang, Yuan, Zhao, Yang, Gu, Wu, Chen, Qiao, Long, et~al.]{luo2025large}
Junyu Luo, Weizhi Zhang, Ye~Yuan, Yusheng Zhao, Junwei Yang, Yiyang Gu, Bohan Wu, Binqi Chen, Ziyue Qiao, Qingqing Long, et~al.
\newblock Large language model agent: A survey on methodology, applications and challenges.
\newblock \emph{arXiv preprint arXiv:2503.21460}, 2025.

\bibitem[McMahan et~al.(2017)McMahan, Moore, Ramage, Hampson, and y~Arcas]{mcmahan2017communication}
H.~Brendan McMahan, Eider Moore, Daniel Ramage, Seth Hampson, and Blaise~Aguera y~Arcas.
\newblock Communication-efficient learning of deep networks from decentralized data.
\newblock In \emph{Proceedings of the 20th International Conference on Artificial Intelligence and Statistics (AISTATS)}, 2017.

\bibitem[Mei et~al.(2024)Mei, Zhu, Xu, Hua, Jin, Li, Xu, Ye, Ge, and Zhang]{mei2024aios}
Kai Mei, Xi~Zhu, Wujiang Xu, Wenyue Hua, Mingyu Jin, Zelong Li, Shuyuan Xu, Ruosong Ye, Yingqiang Ge, and Yongfeng Zhang.
\newblock Aios: Llm agent operating system.
\newblock \emph{arXiv preprint arXiv:2403.16971}, 2024.

\bibitem[{OpenAI}(2025)]{openai_api_2025}
{OpenAI}.
\newblock Openai api documentation, 2025.
\newblock URL \url{https://openai.com/api/}.
\newblock Accessed: 2025-05-16.

\bibitem[Qiu et~al.(2024)Qiu, Lam, Li, Acharya, Wong, Darzi, Yuan, and Topol]{qiu2024llm}
Jianing Qiu, Kyle Lam, Guohao Li, Amish Acharya, Tien~Yin Wong, Ara Darzi, Wu~Yuan, and Eric~J Topol.
\newblock Llm-based agentic systems in medicine and healthcare.
\newblock \emph{Nature Machine Intelligence}, 6\penalty0 (12):\penalty0 1418--1420, 2024.

\bibitem[Wang et~al.(2023)Wang, Xie, Jiang, Mandlekar, Xiao, Zhu, Fan, and Anandkumar]{wang2023voyager}
Guanzhi Wang, Yuqi Xie, Yunfan Jiang, Ajay Mandlekar, Chaowei Xiao, Yuke Zhu, Linxi Fan, and Anima Anandkumar.
\newblock Voyager: An open-ended embodied agent with large language models.
\newblock \emph{arXiv preprint arXiv:2305.16291}, 2023.

\bibitem[Wu et~al.()Wu, Bansal, Zhang, Wu, Li, Zhu, Jiang, Zhang, Zhang, Liu, et~al.]{wu2024autogen}
Qingyun Wu, Gagan Bansal, Jieyu Zhang, Yiran Wu, Beibin Li, Erkang Zhu, Li~Jiang, Xiaoyun Zhang, Shaokun Zhang, Jiale Liu, et~al.
\newblock Autogen: Enabling next-gen llm applications via multi-agent conversation.
\newblock In \emph{ICLR 2024 Workshop on Large Language Model (LLM) Agents}.

\end{thebibliography}
\bibliographystyle{iclr2026_conference}
\newpage
\appendix

\end{document}